\documentclass[10pt]{article} 
\usepackage[preprint]{tmlr}

\usepackage{amsmath,amsfonts,bm}

\def\eqref#1{equation~\ref{#1}}

\def\1{\bm{1}}

\DeclareMathAlphabet{\mathsfit}{\encodingdefault}{\sfdefault}{m}{sl}
\SetMathAlphabet{\mathsfit}{bold}{\encodingdefault}{\sfdefault}{bx}{n}

\usepackage{graphicx}
\usepackage{hyperref}
\usepackage{subfig}
\usepackage{url}

\newcommand{\hide}[1]{}

\title{Generating Human Understandable Explanations for Node Embeddings}

\author{\name Zohair Shafi \email shafi.z@northeastern.edu \\
        \addr Northeastern University \\
        Bostto those of, MA, USA 
	\AND
	\name Ayan Chatterjee \email chatterjee.ay@northeastern.edu  \\
    \addr Northeastern University \\
    Boston, MA, USA 
    \AND
    \name Tina Eliassi-Rad \email t.eliassirad@northeastern.edu \\
    \addr Northeastern University \\
    Boston, MA, USA 
}

\def\month{MM}  
\def\year{YYYY} 
\def\openreview{\url{https://openreview.net/forum?id=XXXX}} 

\begin{document}

\maketitle

\begin{abstract}

Node embedding algorithms produce low-dimensional latent representations of nodes in a graph. These embeddings are often used for downstream tasks, such as node classification and link prediction. In this paper, we investigate the following two questions: (Q1) Can we explain each embedding dimension with human-understandable graph features (e.g. degree, clustering coefficient and PageRank). (Q2) How can we modify existing node embedding algorithms to produce embeddings that can be easily explained by human-understandable graph features? We find that the answer to Q1 is yes and introduce a new framework called \textsc{XM} (short for \textsc{eXplain eMbedding}) to answer Q2. A key aspect of \textsc{XM} involves minimizing the nuclear norm of the generated explanations. We show that by minimizing the nuclear norm, we minimize the lower bound on the entropy of the generated explanations. We test \textsc{XM} on a variety of real-world graphs and show that \textsc{XM} not only preserves the performance of existing node embedding methods, but also enhances their explainability.

\end{abstract}

\section{Introduction}

Graph datasets are ubiquitous from social, to information, to physical, to biological networks. The abundance of graph data has inspired the development of models and algorithms in the field of graph machine learning~\citep{chami2022machine}, a popular task being node embedding,\footnote{Sometimes node embedding is referred to as graph embedding.} where nodes are embedded in a low-dimensional latent space. Many algorithms exist for embedding a graph's nodes into a low-dimensional space. ~\cite{zhang2021systematic} compared a variety of such algorithms on tasks such as greedy routing and link prediction. Recently, much attention has been devoted to explainable graph machine learning~\citep{burkart2021survey}, in part because the performance of such models can often be misleading~\citep{sesh2022sdm}, \citep{sesh2024pnas}. Most works on explainable graph machine learning focus on explaining predictions for downstream tasks. In this work, we study how to assign human-understandable features to node embeddings. For example, a human-understandable embedding dimension represents nodes with many neighbors and few triangles. We also discuss why the objective functions of common node embedding algorithms are not designed to produce human-understandable node embeddings, and how they can be extended to produce such desirable node embeddings.

Recent work on explaining graph machine learning revolves around explaining decisions made on downstream tasks~\citep{pope2019explainability}, \citep{baldassarre2019explainability}, \citep{ying2019gnnexplainer}, \citep{luo2020parameterized}, \citep{vu2020pgm}, \cite{yuan2021explainability}, \citep{liu2018interpretation}]. An outlier here is \cite{gogoglou2019interpretability}'s work. They investigate individual dimensions of an embedding and define an interpretability score for each dimension to measure how well each dimension defines a subgroup of nodes (see Section \ref{related_work} for details). 
In contrast, our aim is to explain each embedding dimension in terms of a user-defined set of human-understandable (a.k.a.~``sense'') features. A byproduct of our approach is that for each node in the graph we can explain its position in the embedding space. Given that many embedding algorithms exist, we propose a framework (called \textsc{XM}) to extend these existing algorithms to make their embeddings explainable. The predictive performance and runtime results of the \textsc{XM} variants are comparable to those of the original methods with AUC on link prediction being within 2\% of the original methods on average across all datasets.

Concretely, we present our investigation of the following two questions. \textbf{Q1: Can we explain dimensions of a node embedding method with human-understandable graph features?} Examples of such features are degree, clustering coefficient, eccentricity, and PageRank. These features can be any set of features that a practitioner decides to use and are not limited to graph features. ~\cite{Henderson2012} called such human-understandable features ``sense'' features. We will borrow their terminology in this paper. \textbf{Q2: Can we modify existing node embedding algorithms to produce embeddings that are explained with sense features? If so, how?} We present a new framework called \textsc{XM} (short for \textsc{eXplain eMbedding}) to answer this question. \textsc{XM} adds two constraints to an existing objective function for node embedding: (a) sparsity w.r.t.~explaining an embedding dimension and (b) orthogonality between embedding dimensions. We provide an ablation study to analyze the impact of each constraint. \textsc{XM} outputs explainable node embeddings that are tested on a variety of real-world graphs, achieving downstream task performances comparable to the state of the art. 

Our main contributions are as follows: 
\begin{itemize}
    \item We present a method for understanding what each dimension of a node embedding method means, with respect to a set of human-understandable sense features defined on nodes. This helps us to understand the placement of a node in the embedding space. These explanations are independent of any downstream task and are in the form of an \emph{Explain} matrix, whose rows represent embedding dimensions and whose columns represent sense features.

    \item To evaluate Explain matrices, we use their nuclear norms and show that minimizing the nuclear norm of an Explain matrix removes noise and reduces the lower bound of its entropy, which is a desirable property. 
    
    \item We introduce the \textsc{XM} framework, which modifies the objective function of any existing node embedding method to produce embeddings that provide better quality explanations. We analyze the impact of each of \textsc{XM}'s constraints (sparsity and orthogonality) through an ablation study and show that using both constraints leads to the largest reduction in nuclear norms.  We demonstrate the effectiveness of \textsc{XM} by modifying a number of embedding algorithms on different real-world datasets and evaluating them on the downstream task of link prediction.
\end{itemize}

\section{Explaining embeddings}
\subsection{Sense-making}
\label{sense_making_section}
In order to explain each embedding dimension, we define a set of human-understandable node features, henceforth referred to as ``sense'' features. For this work, we use global and local properties of the network as our ``sense'' features. Following ~\cite{Ghasemian2020}, we look at the following 15 features: degree, weighted degree (if the edges are weighted), clustering coefficient \citep{network_science_barabasi}, average of the personalized PageRank vector, standard deviation of the personalized PageRank vector,  average degree of the neighboring nodes,  average clustering coefficient of neighboring nodes, number of edges in the ego net, structural hole constraint \citep{burt2004structural}, betweenness centrality \citep{freeman1977set}, eccentricity, PageRank, degree centrality, Katz centrality \citep{katz1953new} and eigenvector centrality \citep{newman2010networks}. Since some of these features are highly correlated, we choose to focus on the following subset of them throughout this work: degree, clustering coefficient, standard deviation of the personalized PageRank, average degree of the neighboring nodes, average clustering coefficient of neighboring nodes, eccentricity and Katz centrality. 

Given an embedding vector for a node $k$, $\vec{y}_k \in \mathbb{R}^{d \times 1}$ and its corresponding sense feature vector $\vec{f}_k \in \mathbb{R}^{f \times 1}$ (where $d$ is the number of dimensions of the embedding space and $f$ is the number of sense features used, each being normalized between 0 and 1), we compute the element-wise similarity between the embedding vector for a node and its corresponding sense feature vector as follows:
\begin{equation}
\begin{aligned}[b]
    \label{explain}
    E_k &= \frac{\vec{y}_k \otimes \vec{f}_k^T}{\|\vec{y}_k\| \|\vec{f}_k\|}
\end{aligned}
\end{equation}

We call $E \in \mathbb{R}^{d \times f}$ the \textit{Explain} matrix (we drop the subscript $k$ corresponding to node $k$, for brevity). Normalized to a range of 0 and 1, each row $i$ of the Explain matrix corresponds to a dimension in the embedding space and each column $j$ corresponds to a sense feature. The value $E_{ij}$ corresponds to how much the dimension $i$ is defined by the sense feature $j$. This matrix $E$, helps us to investigate the linear relationships between the sense feature vectors and the embedding vector.

\subsubsection*{Structural vs Positional Sense Features}
The set of features defined above is not exhaustive and was chosen for uniformity in various data sets. In cases where datasets have inherent node features, we ignore them and use the features defined above for a fair comparison across datasets. However, the choice of sense features is important, and we discuss the impact of different types of features in this section with toy examples.

Observe how the sense features defined above are all structural, meaning that they do not contain any information about the location or position of a node in the network. The sense feature vectors would be similar for two nodes with similar neighborhood structures in the same network. The explanations produced by these sense features would therefore be more role-based, i.e. would speak more to the role of a node in the network (e.g. gatekeeper node or hubs).

\begin{figure}[h]
    \begin{center}
    \includegraphics[width=1.0\linewidth]{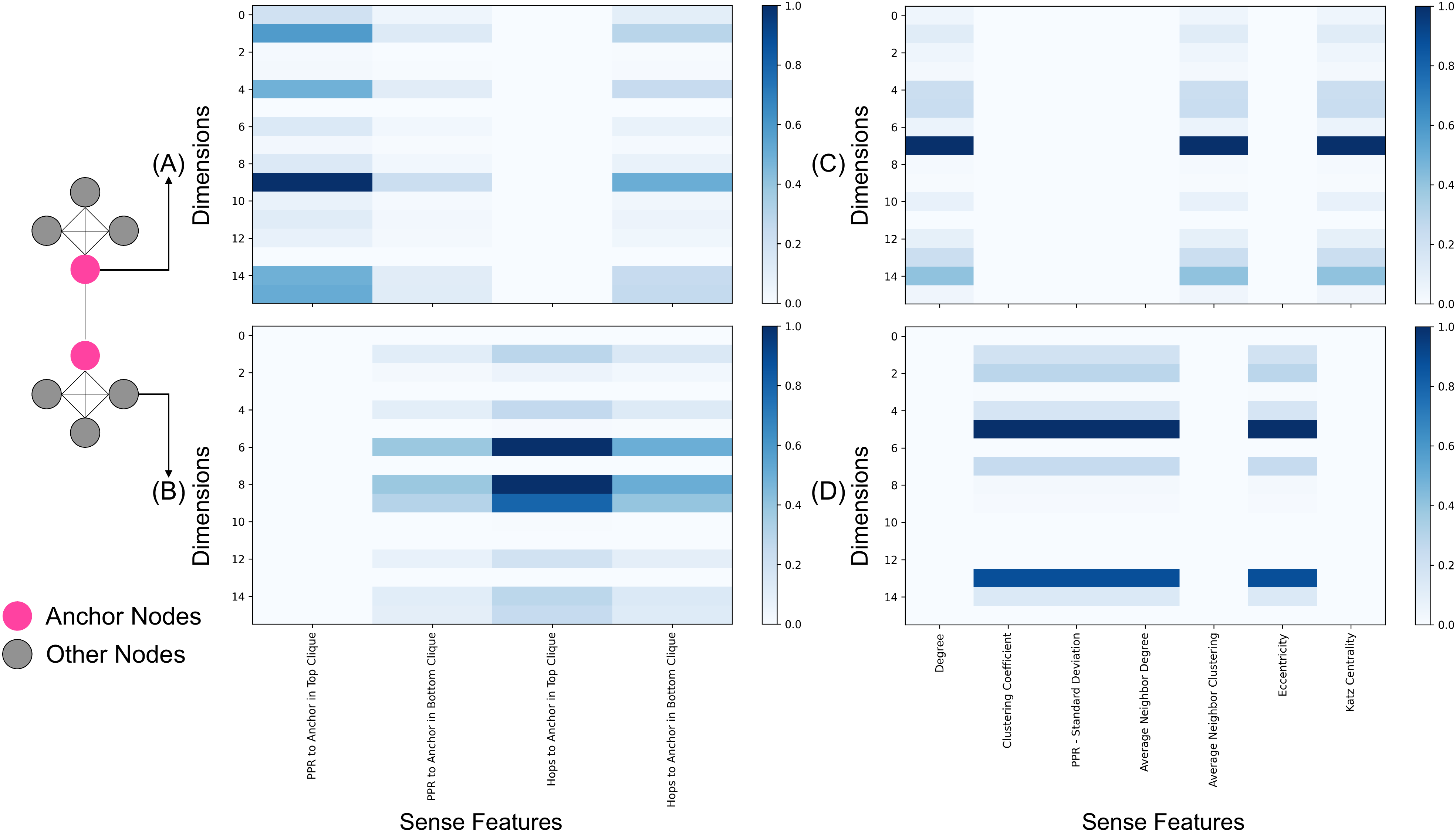}
    \end{center}
    \caption{We show the sense making process using a simple toy example of a barbell graph. We embed the graph into 16 dimensions using DGI and show the Explain matrices for a bridge node (shown in pink) and a random node in a clique (shown in grey). The columns ($x$ axis) of the Explain matrix correspond to the sense features and the rows ($y$ axis) represent each embedding dimension. We demonstrate the utility of these Explain matrices and the differences seen when using different sets of sense features. (A)-(B) Explain matrices for the pointed nodes using positional sense features (shown on the $x$ axis). (C)-(D) Explain matrices for the same two nodes, but using the structural sense features (shown on the $x$ axis). Observe sub plot (C) where degree and average neighbor clustering stand out for the bridge nodes (green) and sub plot (D) where features like clustering coefficient and average neighbor degree stand out for the other nodes (grey). In the case of the structural sense features, all grey nodes would have similar Explain matrices given that they would all have similar sense features (i.e. degree, clustering coefficient etc..), but if passed in positional sense features instead, the grey nodes in different cliques would have different Explain matrices, as seen in sub plot (B) where the feature "Hops To Anchor In Bottom Clique" stands out.} 
    \label{fig:figure_1}
\end{figure}

However, if the explanations one is looking for are more positional in nature,\footnote{Positional explanations look at the role of a node instead of its community.} structural sense features might not be very helpful. To demonstrate the differences in sense features, as well as what the Explain matrix contains, we use a simple barbell graph as a toy example (shown in Figure \ref{fig:figure_1}). We define a secondary set of positional sense features. We follow \cite{you2019position} in defining the concept of anchor nodes and use simple features such as the number of hops to the anchor node and personalized PageRank of the anchor nodes as our positional sense features. We embed the graph using DGI and generate the Explain matrices for each node in the graph following Equation (\ref{explain}). Figures \ref{fig:figure_1}(A) and (B) show the Explain matrices for the bridge node in the top clique (pink) and a random node in the bottom clique (gray) using our simple positional features, while Figures \ref{fig:figure_1}(C) and (D) show the Explain matrices for the same nodes, but using the structural sense features described above. Observe in Figures \ref{fig:figure_1}(C) and (D) how degree and average neighbor clustering stand out for the bridge nodes (shown in pink) and features such as the clustering coefficient and average neighbor degree stand out for the other nodes (shown in gray). Note that in case of using structural sense features, all gray nodes (non-bridge nodes) would have similar sense features and embeddings; and therefore, similar Explain matrices, but, in the case of positional sense features (Figure \ref{fig:figure_1}(B)), the gray nodes in different cliques would have different sense feature vectors leading to different Explain matrices. (Observe that the feature "Hops To Anchor In Bottom Clique" stands out in Figure \ref{fig:figure_1}(B)).

Based on this toy example, it is clear that the set of 7 structural sense features used is not exhaustive. We use these 7 features for consistency across datasets but encourage researchers to tailor the chosen sense features to the desired type of explanations.

\subsubsection{Real world example}
\begin{figure}[h]
    \begin{center}
        \includegraphics[width=1.0\linewidth]{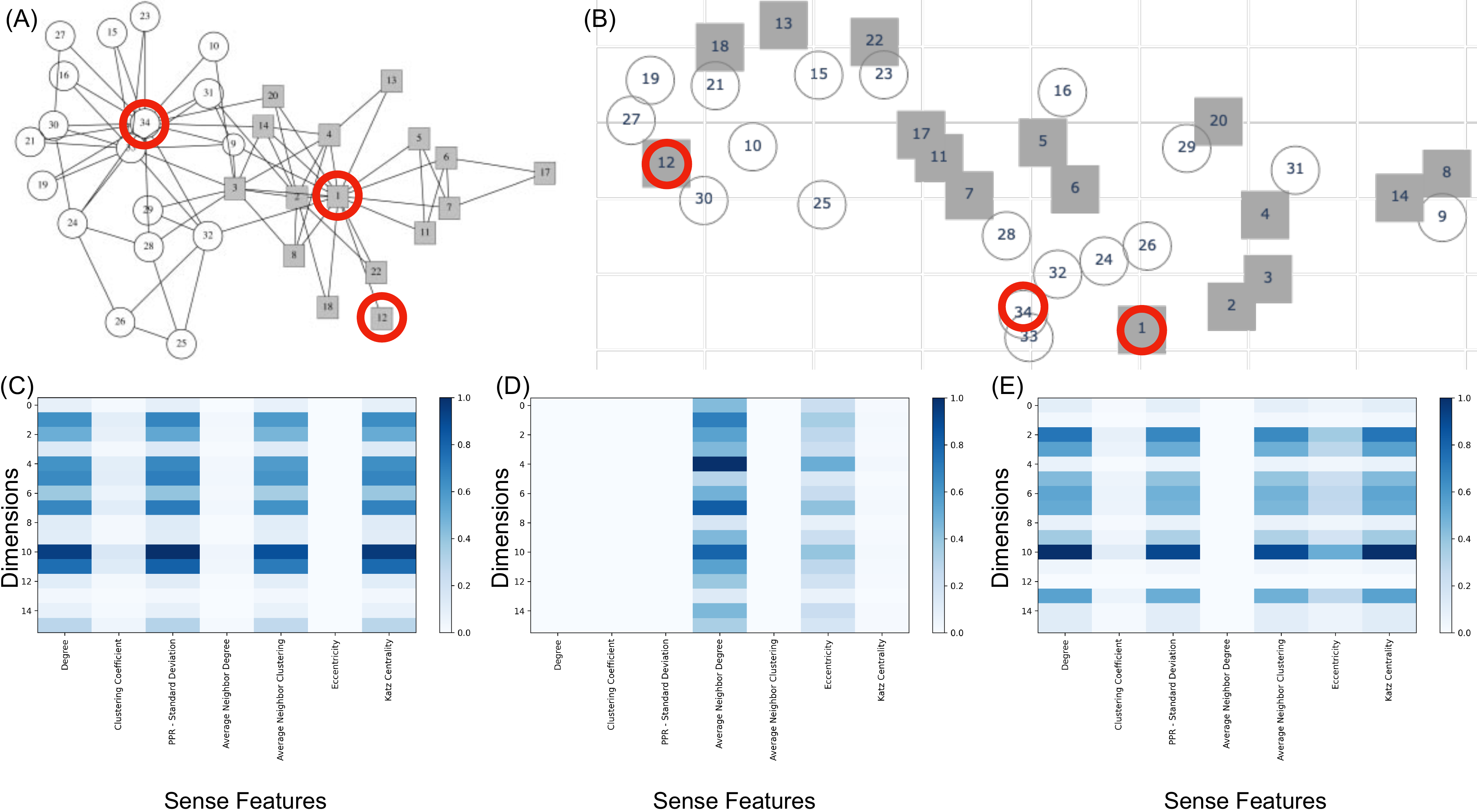}
    \end{center}
    \caption{\textbf{(A)} Karate Club Network~\citep{karate}. \textbf{(B)} We embed the Karate Club network into 16 dimensions using DGI and visualize it in 2 dimensions using UMAP~\citep{mcinnes2018umap}. Explain matrix for \textbf{(C)} the instructor (node 1). \textbf{(D)} Explain matrix for a random student node (node 12). \textbf{(E)} Explain matrix for the president (node 34). Examining the Explain matrices for each of these nodes gives us an understanding of their placement in the embeddings space. Observe how features such as degree, personalized page rank etc., stand out for nodes 1 and 34 (subplots \textbf{(C)} and \textbf{(E)}) and features like average neighbor degree stands out for the random student node (subplot \textbf{(D)}).}
    \label{fig:figure_2}
\end{figure}

 We look at the Karate Club Network for a more concrete real-world example. We chose this network for brevity and pedagogy and discuss larger networks in the Results section. Figure \ref{fig:figure_2}(A) shows the original Karate Club Network~\citep{karate}. We embed the Karate Club graph into 16 dimensions using DGI. We then visualize these embeddings by projecting them into 2 dimensions using UMAP \citep{mcinnes2018umap} (shown in Figure \ref{fig:figure_2}(B)). For a detailed node-wise view, we look at the president of the club (node 34), the instructor (node 1), and a random student (node 12). We use Equation (\ref{explain}) to compute the Explain matrix for each node. Figure \ref{fig:figure_2}(C) corresponds to the Explain matrix for the instructor (node 1), Figure \ref{fig:figure_2}(D) corresponds to the random student (node 12), and Figure \ref{fig:figure_2}(E) to the president (node 34). Examining the Explain matrices for nodes 1 and 34 (Figures \ref{fig:figure_2}(C) and (D)), we see similar sense features (like degree, the standard deviation of the personalized PageRank, Katz centrality and average neighbor clustering) stand out and are explained most by dimensions 14 and 15 (note that the specific dimensions are not constant). Also, observe that these two nodes are placed close together in the embedding space (low-dimensional visualization in Figure \ref{fig:figure_2}(B)). We also see that the Explain matrix for node 12 (Figure \ref{fig:figure_2}(D)) is  different, and has average neighbor degree and eccentricity as its distinguishing features. 

\subsection{Our Proposed Framework: \textsc{XM} (\textsc{eXplain eMbedding})}
\label{xm_section}
\subsubsection*{Quantifying the Quality of Explanations}
Given the method for generating explanations, we quantify the quality of these explanations. The (potentially) large number of dimensions and sense features drives us to use heatmaps to visualize their relationships. Figure \ref{fig:figure_2} shows the heatmap visualization of the Explain matrix. Notice that a noisy Explain matrix is hard to interpret, but sparser Explain matrices with each dimension being defined by unique sets of sense features are easier to interpret. 

Since what we are looking for is a denoised Explain matrix, we follow the work from image denoising using nuclear norm minimization [\cite{gu2014weighted}, \cite{xu2017multi}]. We report the nuclear norms of the Explain matrices as a quantitative metric, with lower nuclear norms leading to better explanations.

\subsubsection{Augmenting Embeddings}

Note that we do not want a perfect one-to-one mapping between the embedding dimensions and the sense features. If a one-to-one correspondence is imposed, then the network information beyond that contained in the simple node sense features is lost while creating the embeddings. Such embeddings would not capture higher-order network properties that contribute to downstream tasks \citep{Ghasemian2020}, resulting in poor performance. Using sense features themselves as embedding vectors or performing dimensionality reduction on a large set of sense features suffers from the same issue of not capturing higher-order network properties. 
To that end, we propose \textsc{XM}, a framework for augmenting existing embedding algorithms that includes sense feature information, as well as information that the underlying algorithm captures.

\textsc{XM} incorporates sense features and achieves a less noisy Explain matrix by imposing a sparsity constraint along the columns of the Explain matrix. This aims to zero-out features with small contributions to the definition of a dimension. Following recent advances in dimensional contrastive learning by \cite{nguyen2023dimcl}, XM also imposes an orthogonality constraint along the rows of the Explain matrix which aims to have embedding dimensions that are defined by different sets of sense features. XM uses the two constraints, sparsity and orthogonality, instead of directly minimizing the nuclear norm for added control over the objective function (see section \ref{ablation}). The loss terms for the two constraints can be written as follows. 
\begin{align}
    \label{sparse_eq}
    L_{sparse} &= \|E_k[:, j]\|_1 & \forall j \in columns &
\end{align}  
\begin{align}
    \label{ortho_eq}
    L_{ortho} &= \|E_k[i, :] E_k^T[j, :]\|_2 &\forall i \in rows\ & \forall j \in columns
\end{align}

These two additional loss terms are then added to the objective functions of existing embedding algorithms. As an example, the objective for SDNE consists of 3 loss terms - a first order loss term, a reconstruction error term and a regularization term:
\begin{align}
    L_{SDNE} = \alpha L_{first} + \beta L_{recon} + \nu L_{reg}
\end{align}
where $\alpha$, $\beta$, and $\nu$ are tunable hyperparameters.

We augment this objective as follows : 
\begin{align}
L_{SDNE} = \alpha L_1 + \beta L_2 + \nu L_{reg} + \gamma L_{Sparse} + \delta L_{Ortho}  
\end{align}
which we refer to as SDNE+XM. Here, $\gamma$ and $\delta$ are hyperparameters for the sparsity and orthogonality constraints, respectively.

In order to demonstrate our method, we apply our augmentation to 4 unsupervised embedding algorithms across 6 different networks. We look at the taxonomy of machine learning on graphs by \cite{chami2022machine} and pick two algorithms from the branch where $X=I$, (i.e. these algorithms do not incorporate node attributes) and two algorithms from the branch where $X \neq I$ (i.e. these algorithms do allow node attributes). We choose these algorithms because providing sense features as node attributes is a simple and straightforward method to incorporate the sense features into the embeddings. However, as we show in the Results section, we achieve lower nuclear norms by augmenting the objective function.

Table \ref{table:algo_table} shows the algorithms selected and Table \ref{table:real_nets} summarizes the datasets used. These have been chosen to span a range of average degrees and clustering coefficients. 

\begin{table}[t]
    \caption{Embedding algorithms used. Classification of these algorithms was adopted from Chami et al.~\cite{chami2022machine}, where we pick two algorithms that allow for a node feature matrix, and two algorithms that do not. Providing sense features as node attributes is a simple method of incorporating the sense features into the embeddings. XM variants outperform the original versions in terms of nuclear norms.}
    \begin{center}
        
    \begin{tabular}{cccl}
    Algorithm & Node Features & Sub Type \\
    \hline
    SDNE \cite{Wang2016} & No ($X=I$) & Autoencoder \\
    LINE \cite{tang2015line} & No ($X=I$) & Skip Gram \\
    DGI \cite{velickovic2019deep} & Yes ($X \neq I$) & Message Passing \\
    GMI \cite{peng2020graph} & Yes ($X \neq I$) & Message Passing \\
    \end{tabular}
    \end{center}
    \label{table:algo_table}
\end{table}

\begin{table}[t]
    \caption{Real-world networks used in our experiments. $\langle k \rangle$ refers to the average degree, $\sigma_k$ refers to the standard deviation of the degree, $r$ refers to the degree assortativity and $c$ refers to the average clustering coefficient. These networks were picked to span a range of average degrees and clustering coefficients.}
    \begin{center}
    \begin{tabular}{ccccccccl}
    Network & Nodes & Edges & $\langle k \rangle$ & $\sigma_k$ & $r$ & $c$ & Transitivity\\
    \hline
    EU Email \cite{leskovec2007graph} & 986 & 16687 & 33.84 & 37.81 & -0.01 & 0.40 & 0.26 \\ 
    US Airport \cite{zhu2021node} & 1186 & 13597 & 22.92 & 40.49 & 0.03 & 0.50 & 0.42 \\ 
    Squirrel \cite{rozemberczki2021multi} & 5201 & 198493 & 76.32 & 161.45 & -0.23 & 0.42 & 0.34 \\
    Citeseer \cite{bollacker1998citeseer} & 3327 & 4676 & 2.81 & 3.38 & 0.05 & 0.14 & 0.13 \\ 
    FB15K-237 \cite{toutanova2015observed} & 14951 & 261581 & 34.99 & 111.43 & -0.10 & 0.213 & 0.02 \\ 
    PubMed \cite{roberts2001pubmed} & 19717 & 44327 & 4.49 & 7.43 & -0.04 & 0.06 & 0.05 \\
    \end{tabular}
    \end{center}
    \label{table:real_nets}
\end{table}

\subsubsection*{Interpreting Rank Reduction with Nuclear Norms}
 Given a matrix of rank $r$, one can reconstruct it using $r$ rank-1 matrices. The nuclear norm of a matrix is the tightest convex bound of the rank function over the unit ball of matrices with bounded spectral norm.\footnote{Note that the rank of a matrix is a non-convex function, which makes optimization problems involving the rank hard to solve. The nuclear norm serves as a convex proxy for the rank.} Thus, the difference in nuclear norms of two matrices gives us a lower bound on how many more rank-1 matrices are needed to reconstruct the matrices. For example, SDNE on the EU Email dataset (with $d$ = 128) produces an Explain matrix whose nuclear norm is $\simeq 8 \pm 0.97$ (see Figure \ref{fig:figure_7}). However, its Explain matrix has a nuclear norm of $\simeq 1.8 \pm 0.26$ when the sparsity constraint is used, $\simeq 2.1 \pm 0.29$ when the orthogonality constraint is used, and $\simeq 1.6 \pm 0.39$ when both constraints are used. (Note that the error bars throughout denote standard error) This means that the Explain matrix for the EU Email dataset can be approximated with two rank-1 matrices when the sparsity and orthogonality constraints are used, instead of eight rank-1 matrices with no constraints used. 

\subsection*{Nuclear Norm and Matrix Entropy}
For a given node, assume there exists an embedding $y^*$ (we drop the vector notation $\vec{y}$ for brevity), which is the optimal embedding. Let $y^{xm}$ be the embedding for the same node found by \textsc{XM}. Recall that we define the Explain matrix $E$ as:
\begin{align}
    E^* = \frac{y^{*}\otimes f^T}{\|y^*\|\|f\|} \\
    E^{xm} = \frac{y^{xm}\otimes f^T}{\|y^{xm}\|\|f\|}
\end{align}

Let 
\begin{align}
    A = E^{*}E^{*^T} \\
    B = E^{xm}E^{xm^T}
\end{align}

Observe that $A$ and $B$ are now symmetric matrices and therefore have an orthonormal basis of eigenvectors with real eigenvalues. 

We then have 
\begin{align}
    A &= U\Sigma U^T &\text{where } U \text{is orthonormal and } \Sigma \text{ is diagonal} \\
    z^TAz &= z^T\ U\Sigma U^T\ z &\forall z \geq 0 \in \mathbb{R}^n \\
    z^TAz &= \sum_{i = 1}^n \Sigma_{ii} (U^Tz)_i^2 & \\
    z^TAz &\geq 0 &
\end{align}

This shows that $A$ is a positive semi-definite (PSD) matrix. The same proof applies for the matrix $B$. Given that $A$ and $B$ are PSD, \cite{yu2013strong} and \cite{playing_psd} show that when the Von Neumann entropy of a PSD matrix is defined as 
\begin{align}
    H(A) = -tr[A\ log\ A]
\end{align}

and its associated Bregman Divergence is defined as 

\begin{align}
    D(A\|B) = tr[A(log\ A -log\ B)] - tr(A) + tr(B)
\end{align}

the following extension of Pinsker's inequality holds true: 
\begin{align}
    D(A\|B) \geq \frac{1}{2} \|A-B\|_*^2
\end{align}

We also know that the nuclear norm is an upper bound on the $L_2$ norm, i.e. 
\begin{align}
    \|A\|_* \geq \|A\|_2
\end{align}

This gives us
\begin{align}
    D(A\|B) \geq \frac{1}{2} \|A-B\|_*^2 \geq \frac{1}{2} \|A-B\|_2^2 \\
    D(A\|B) \geq \frac{1}{2} [\|A\|_2^2 - 2\|A\|_2\|B\|_2 + \|B\|_2^2]
\end{align}

Recall that 
\begin{align}
    B = E^{xm}E^{xm^T}
\end{align}

which is explicitly minimized as the orthogonality constraint in the \textsc{XM} framework. Notice that the $\|A\|_2$ term is constant since it is the optimal solution. Thus, by minimizing the $L_2$ norm of $B$, we reduce the lower bound in the entropy between an optimal explanation $E^*$ and the explanation $E^{xm}$ found by \textsc{XM}.

\section{Experiments}
\label{results_section}
We start by visualizing the Explain matrices for the Karate Club network embedded using DGI. We then discuss quantitative metrics for the networks from Table \ref{table:real_nets} embedded using each of the 4 embedding algorithms from Table \ref{table:algo_table} and their \textsc{XM} variants.

\subsection{Explanations - Visual Evaluation}
\begin{figure}[h]
    \begin{center}
        \includegraphics[width=1.0\linewidth]{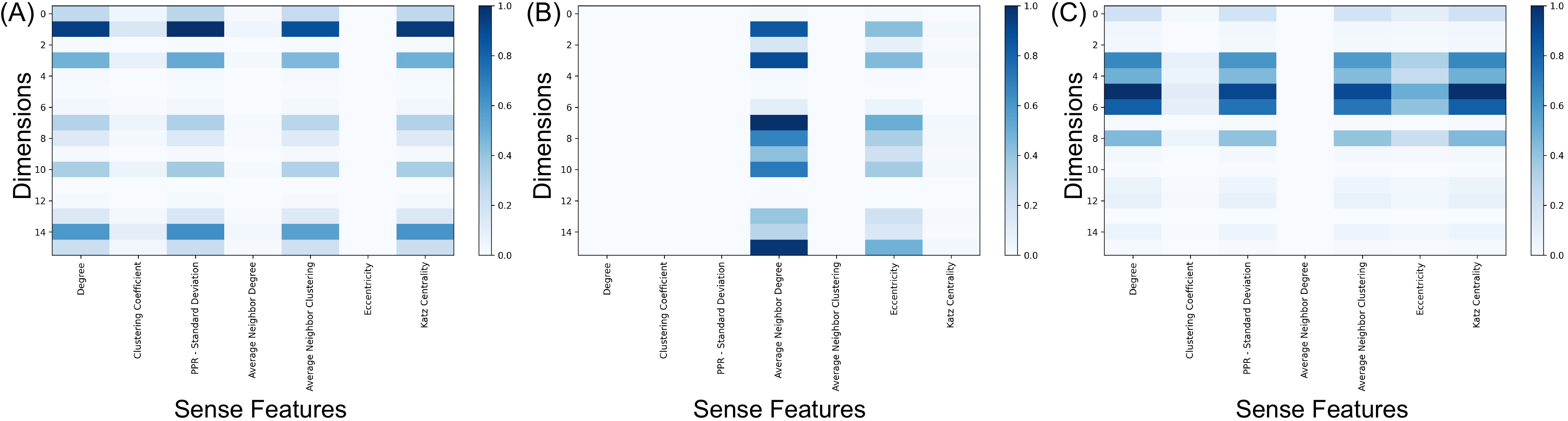}
    \end{center}
    \caption{{Explain matrices for the Karate Club network~\citep{karate} generated using DGI+XM for (A) the instructor (node 1), (B) a random student (node 12), and (C) the president (node 34). Compare each of (A), (B), and (C) to Figure \ref{fig:figure_2}(C),(D), and (E), respectively. Observe that the Explain matrices in this figure (generated by DGI+XM) are sparser and each sense feature is explained by fewer dimensions, when compared to the Explain matrices from Figure \ref{fig:figure_2} (generated by the standard DGI).}}
    \label{fig:figure_3}
\end{figure}

Figure \ref{fig:figure_3}, shows the Explain matrix for the Karate Club network. As in Figure \ref{fig:figure_2} (C)(D) and (E), the three subplots of Figure \ref{fig:figure_3} show the Explain matrices for the instructor node (node 1, Fig \ref{fig:figure_3}(A)), a random student node (node 12, Fig \ref{fig:figure_3}(B)) and the president node (node 34, Fig \ref{fig:figure_3}(C)). To stay consistent across experiments, these plots were generated using 16-dimensional embeddings that were generated from DGI+XM. Observe that the Explain matrices shown in Figure \ref{fig:figure_3}(A)(B) and (C), while highlighting similar sense features, are sparser than the respective plots in Figure \ref{fig:figure_2}(C)(D) and (E) and each sense feature is defined by fewer dimensions, i.e. the dimensions are orthogonal in their explanations when compared to the standard DGI. 

\subsection{Explanations - Quantitative Evaluation}
\begin{figure}[h]
    \begin{center}
        \includegraphics[width=1.0\linewidth]{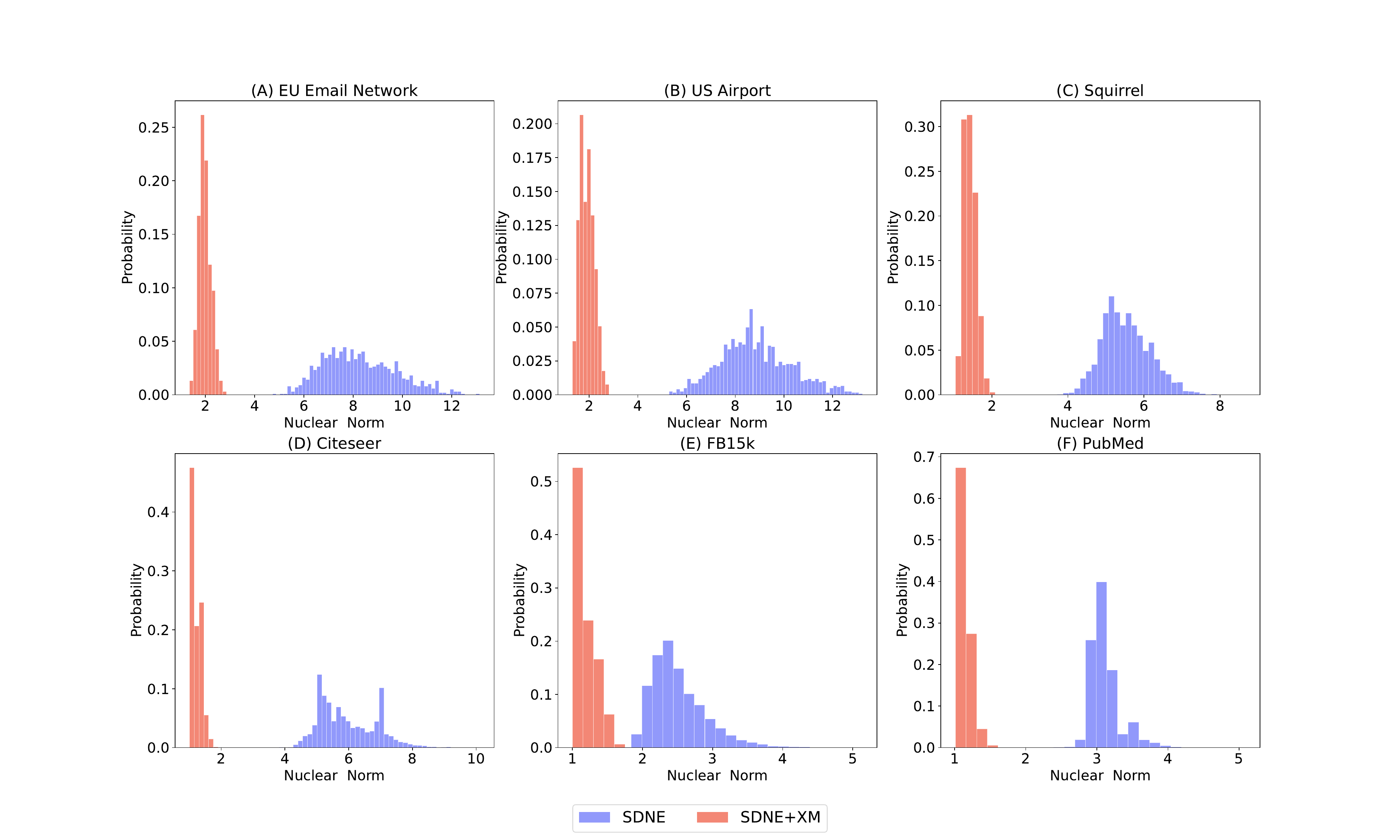}
    \end{center}
    \caption{Distribution of nuclear norms of the Explain matrices for each node for SDNE and SDNE+XM. Observe how the distribution of SDNE+XM is shifted to the left with a lower mean across each of the 6 datasets. See Section~\ref{nuclear_norm_distribution_section} in the appendix for results on the other methods. 
    }
    \label{fig:figure_4}
\end{figure}

We look at the nuclear norms of the Explain matrices as a quantitative metric to evaluate and compare the standard and \textsc{XM} variants of the algorithms. For each dataset, we use each of the 4 algorithms shown in Table \ref{table:algo_table} and compute the mean of the nuclear norm of the Explain matrices across all nodes in the dataset. We embed every dataset into 128 dimensions for consistency across datasets and algorithms and pass in sense features as node attributes when running DGI and GMI. Figure \ref{fig:figure_4} shows the distribution of the nuclear norms of the Explain matrices for SDNE and SDNE+XM with results for the other algorithms shown in the appendix (Figures \ref{fig:figure_8}, \ref{fig:figure_9}, and~\ref{fig:figure_10}). We see that the distribution of the nuclear norm values shift to the left (i.e. lower) across each dataset. 

\subsection{Link Prediction Performance}
By adding additional loss terms to the objective function, we modify the embeddings themselves to generate a less noisy Explain matrix. Doing so seems to have an impact on downstream tasks, although, we still see comparable performance to the original version. We use link prediction as our downstream task and sample an equal number of positive and negative edges from the graph. We use $60\%$ of the edges as training data and the remaining as test data. The embeddings are passed through a simple 2 layer fully connected neural network. We repeat the entire process 3 times to create a three-fold cross validation setup. We use 128-dimensional embeddings across all algorithms and networks for consistency. 

\begin{figure}[h]
    \begin{center}
        \includegraphics[width=0.8\linewidth]{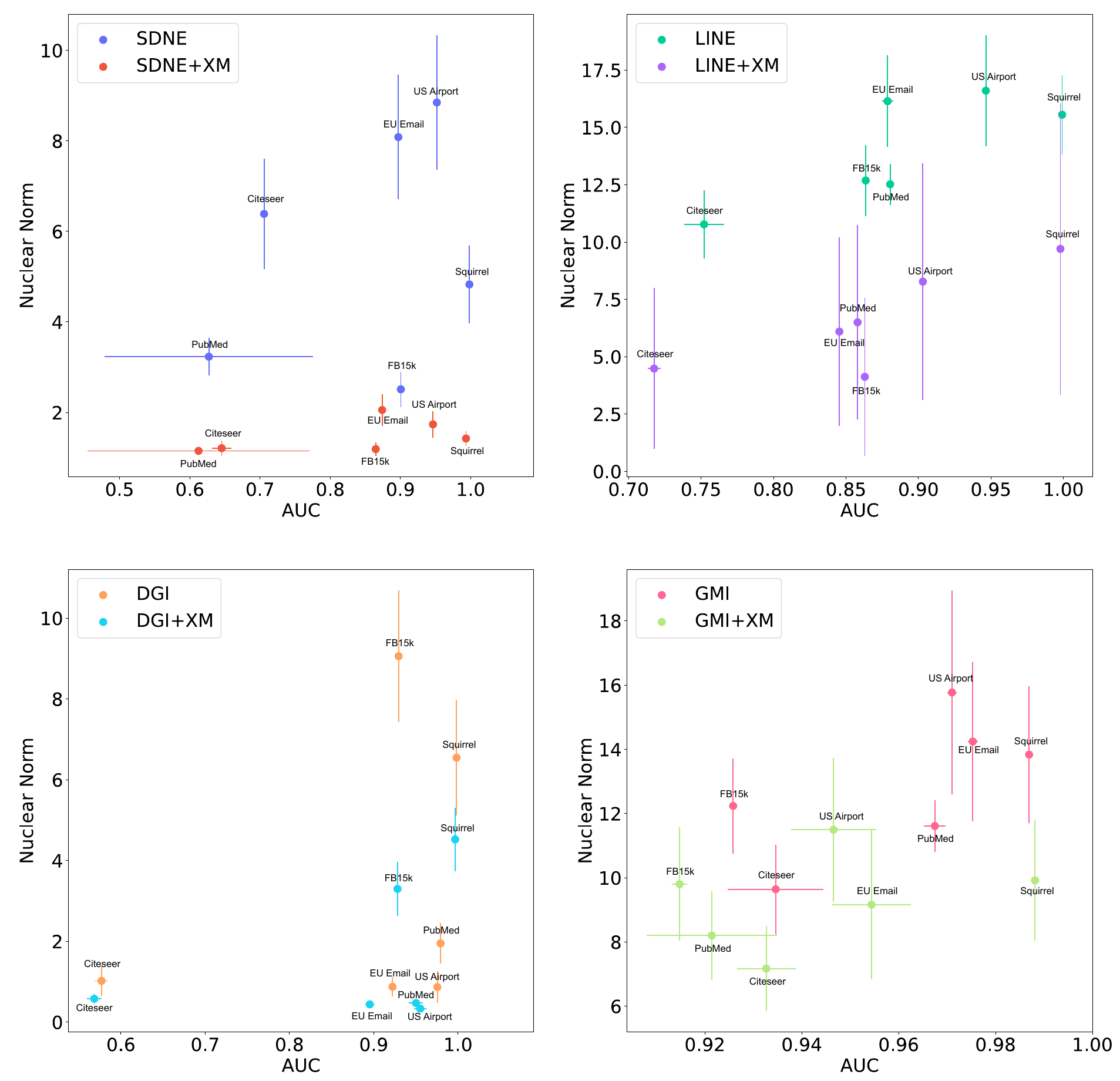}
    \end{center}
    
    \caption{{AUC scores for link prediction across 4 embedding algorithms - SDNE, LINE, DGI, GMI - shown in each subplot, on the 6 networks defined in Table \ref{table:real_nets}. Error bars show standard error. AUC scores for the XM variants are comparable to the original algorithms with nuclear norms of the XM variants being lower. The XM variants also follow the original algorithms in terms of standard error, for example, SDNE has large variance in AUC for the PubMed dataset, which can also be seen for SDNE+XM. Results are from a three-fold cross validation experiment with 128 dimensional embeddings.}}
    \label{fig:figure_5}
\end{figure}

Results are shown in Figure \ref{fig:figure_5} with nuclear norm shown on the $y-$axis and the AUC shown on the $x-$axis with error bars denoting the standard error. We observe that across all algorithms, AUC scores are comparable to the original version with the corresponding nuclear norms being lower. 

\begin{figure}[h]
    \begin{center}
        \includegraphics[width=0.8\linewidth]{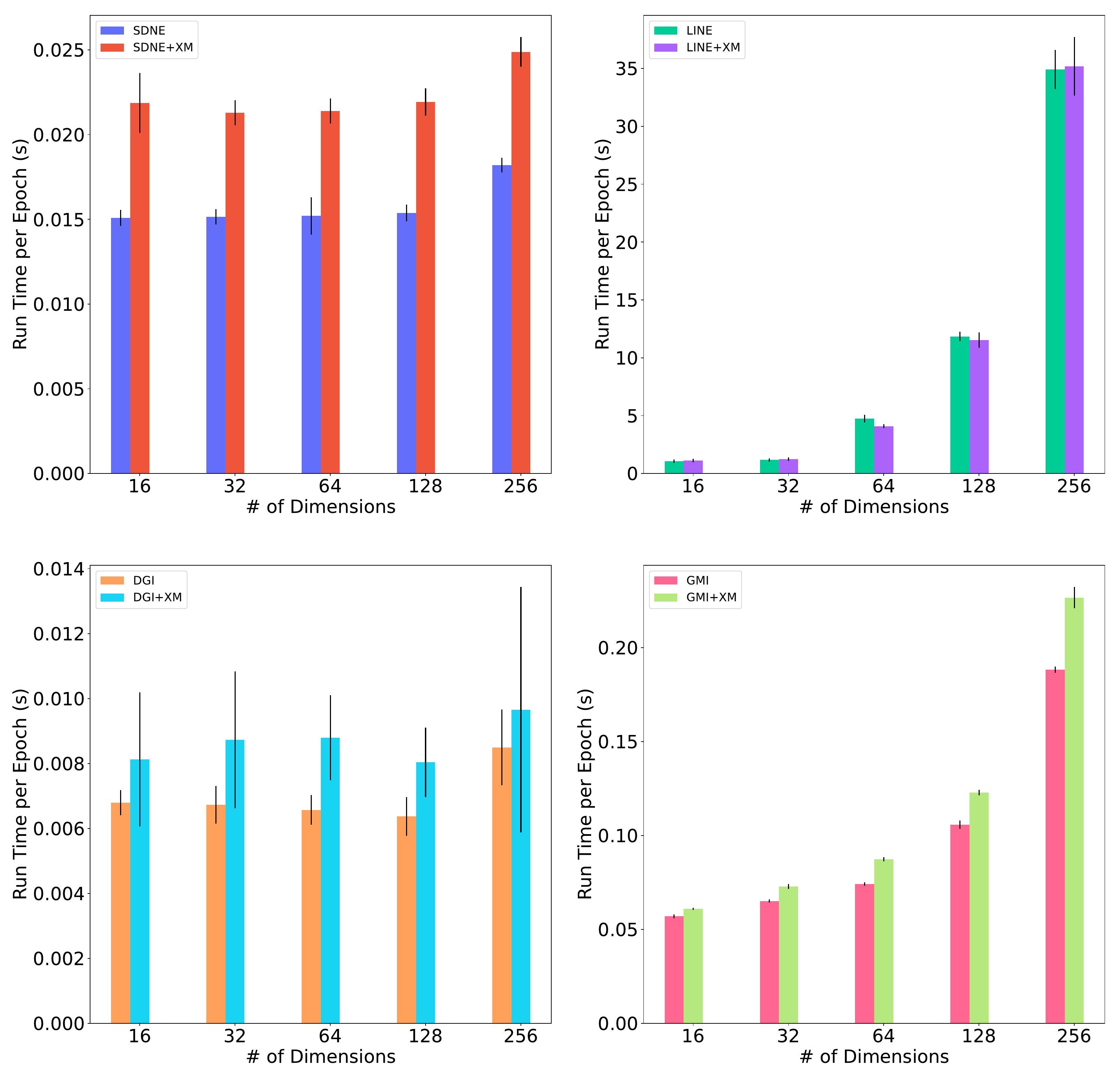}
    \end{center}
    \caption{{EU email network: runtimes per epoch (in seconds) for SDNE, LINE, DGI, GMI, and their corresponding \textsc{XM} variants. Results are averaged across 5 runs. Observe that the runtimes per epoch are comparable to the original versions. Error bars show standard error. See Section~\ref{runtime_analysis_section}} in the Appendix for the resutls on other networks.}
    \label{fig:figure_6}
\end{figure}

We also examine differences in runtime due to the additional constraints in Figure \ref{fig:figure_6}. More specifically, we examine the time (in seconds) per epoch for each of the 4 algorithms we test for the EU Email network. We observe that the \textsc{XM} variants are comparable in runtime to their original versions. Results shown are averaged across 5 runs with error bars showing standard error. Similar run time comparisons for other datasets are shown in the appendix (Figures \ref{fig:figure_11} through~\ref{fig:figure_15}).

\subsection{Ablation Study}
\label{ablation}
\begin{figure}[h!]
    \begin{center}
        \includegraphics[width=1.0\linewidth]{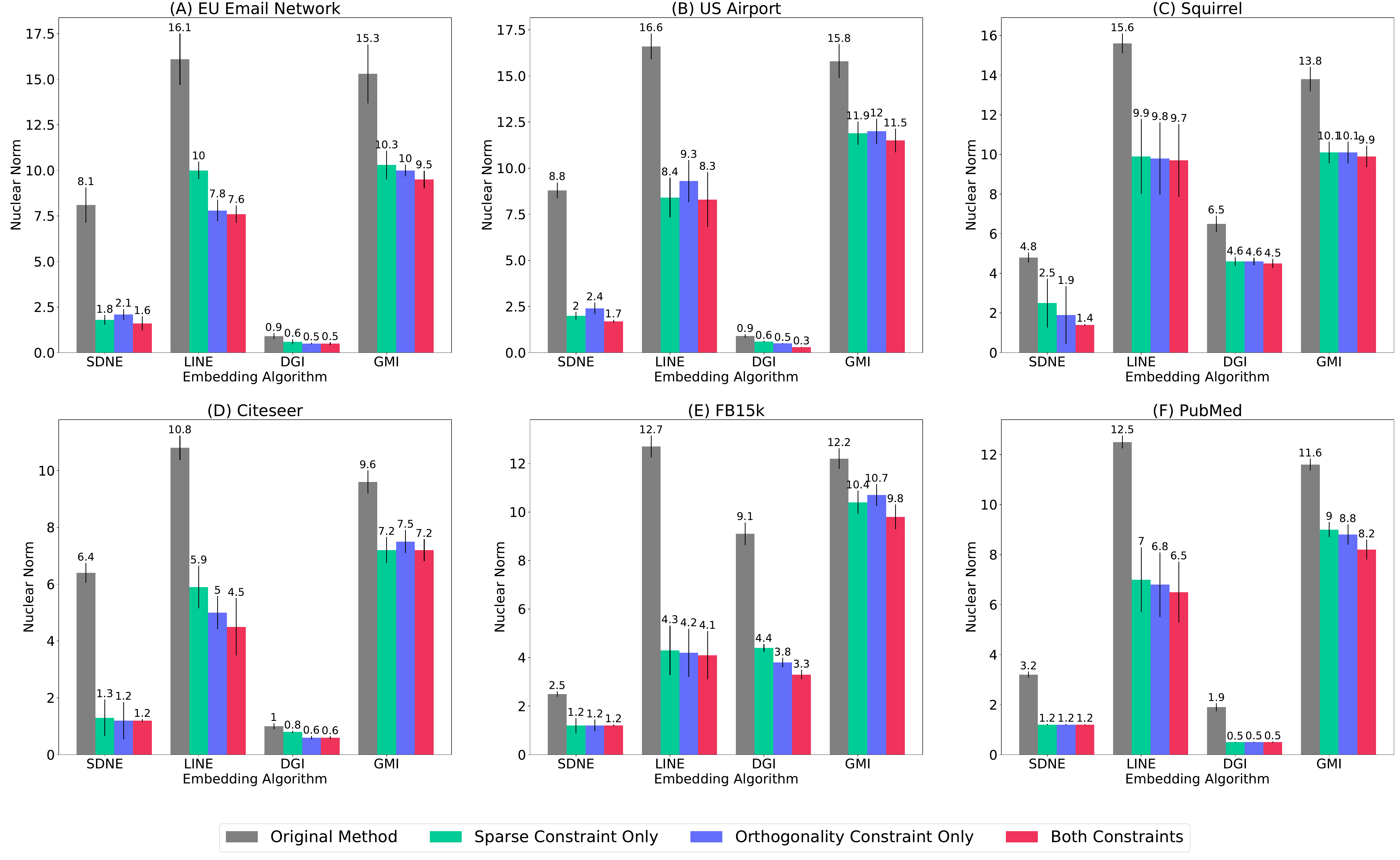}
    \end{center}
    
    \caption{{Ablation study examining the impact of each term in the loss function. Observe that the orthogonality constraint alone reduces nuclear norm more than the sparsity constraint alone. However, using both constraints leads to the lowest nuclear norms. The plot shows average values over 5 runs. Results shown are for 128 dimensional embeddings but hold true across various dimensions. (We tried $d=16, 32, 64, 128, 256$.) We do not find a correlation between the number of embedding dimensions and the reduction in the nuclear norm values. Error bars show standard error. }}
    \label{fig:figure_7}
\end{figure}

We add two constraints, orthogonality and sparsity, while computing the augmented node embeddings. We study how each contributes to the reduction of nuclear norms. Figure \ref{fig:figure_7} shows nuclear norms for each network embedded in 128 dimensions using each of the 4 embeddings algorithms, averaged across 5 runs. We see that the orthogonality constraint alone reduces the nuclear norm more than the sparsity constraint alone. We see that across all algorithms, using both constraints together achieves the lowest nuclear norm. 

\section{Discussion}
\label{limitation_section}

By design, the explanations are granular. They are for each node at the granularity of each dimension. This is useful when inspecting particular nodes, but to get a holistic explanation of a large network will require further analysis of a set of Explain matrices. 

As is common in works on explainability, the use of the \textsc{XM} framework leads to a marginal reduction in performance scores for downstream tasks. However, note that the reduction in performance can be controlled through the choice of hyperparameters. For the this study, we chose to prioritize the quantitative value of the nuclear norm to get the largest reduction possible. In actual use, the practitioner can balance the amount of denoising of the Explain matrices to the desired level of performance in downstream tasks by setting the hyperparameters accordingly. (Our choices for the hyperparameters are available in our online code repository.) 

The usability of \textsc{XM} depends on the choice of sense features provided. A poor choice could lead to Explain matrices that do not contain any useful information. The decision to use structural sense features throughout this work was to provide a set of features that are reasonable and understandable to humans in the graph context (e.g.~node degree and clustering coefficient) and can be used in networks from different domains (e.g.~sociology and biology).

\section{Related Work}
\label{related_work}

Interest in explainable machine learning models has attracted much attention in recent years, in part because the performance of such models can often be misleading~\citep{burkart2021survey}. With the increasing application of machine learning to graphs, the development of explainable models is becoming a necessity. However, most of the work on explainable models focuses on explaining predictions for downstream tasks. For example, ~\cite{pope2019explainability} and ~\cite{baldassarre2019explainability} extend the concept of explainable predictions in images to network data using gradient-based approaches to visualize and understand the predictions of Graph Convolutional Networks (GCNs). 

Another popular set of methods focuses on finding a subgraph that contributes to a particular classification. For example, ~\cite{ying2019gnnexplainer} introduce  \emph{GNNExplainer}, which attempts to find a subgraph and a subset of the node features by maximizing the mutual information between the original prediction and the prediction about a substructure. However, the explanations generated are customized for each instance, which makes them hard to generate. ~\cite{luo2020parameterized} extend~\emph{GNNExplainer} by introducing \emph{PGExplainer}, which utilizes a generative probabilistic model for graph data that has been shown to be able to learn underlying structures from observed data. It uses these underlying structures as explanations and models the underlying structure as edge distributions, where the explanatory graph is sampled. This generative model in \emph{PGExplainer} is parameterized with a deep neural network to collectively explain the predictions of multiple instances. \cite{spinelli2022meta} propose a meta-explainer to improve the quality of explanations during training time. Roughly related to our approach, they aim to steer the optimization procedure towards minima that allow post-hoc explainers like \emph{GNNExplainer} and \emph{PGExplainer} to achieve better results. ~\cite{vu2020pgm} use probabilistic graphical models to explain substructures in a network, and ~\cite{yuan2021explainability} use Monte Carlo tree search to search all possible subgraphs and identify important substructures.

\cite{zhang2021relex} propose RelEx, which identifies important nodes and links in the prediction of a given node, but does so by treating the underlying model as a black box and learning relational explanations. Explanations come in the form of a mask matrix, which, akin to our work, has sparsity constraints. Along the lines of sparsity, ~\cite{lin2020graph} propose GISST (Graph neural networks Including SparSe inTerpretability), which combines an attention mechanism and sparsity regularization to yield an important subgraph and node feature subset related to any graph-based task.~\cite{li2022egnn} use a student-teacher setup to distill knowledge from a large pre-trained GNN to a shallower GNN, while explicitly learning contribution weights between two nodes using an attention mechanism. \cite{schnake2021higher} propose GNN-LRP that uses layer-wise relevance propagation that outputs a collection of walks that are relevant for a given prediction.

\cite{agarwal2022evaluating} provide a framework titled GraphXAI that delves deeper into the different types of GNN explainers. These explainers can be categorized into three types: perturbation-based, gradient-based, and surrogate-based models. These explainers can often be misleading, as the explanations may rely on a different rationale for the test data compared to the rationale learned during training. GraphXAI provides a benchmark for evaluating these explainers by generating ground-truth explanations using the ShapeGGen graph generator, which is capable of generating ground-truth explanations for homophilic, heterophilic, and attributed graphs. We refer the interested reader to the survey by \cite{li2022survey},  which provides a detailed taxonomy of the various graph explanation methods.

All of these explanatory methods depend on a downstream task such as link prediction or node classification. More closely related to our work (i.e. explanations independent of a downstream task), are works by ~\cite{dalmia2018towards} and ~\cite{bonner2019exploring} who study how good an embedding algorithm is at explaining certain node properties (akin to what we have referred to as sense features). ~\cite{liu2018interpretation} utilize network homophily and hierarchically cluster nodes based on the embeddings to create a taxonomy. Explanations come in the form of these taxonomies. Also related would be the study of word analogies in the natural language domain where one could interpret a word as a node in a graph. ~\cite{ethayarajh2018towards} and ~\cite{allen2019analogies} explore the relationships between word embeddings and analogies, however, these are applicable only to certain categories of word embeddings (negative sampling-based methods). ~\cite{yin2018dimensionality} provide a theoretical understanding of word embeddings and its dimensionality and propose a metric on dissimilarity between word embeddings. However, they focus on embedding algorithms that can be formulated as matrix factorizations.\cite{monti2024a} provide an axiomatic approach for auditing explainers for the task of node classification. More specifically, they define a set of `important' features such that a change in these `important' features would cause a change in the classification result and check if an explainer is able to diagnose if an underlying model is using the `important' feature or not. 

In terms of explanations per dimension, ~\cite{gogoglou2019interpretability} define an interpretability score for each dimension in an node embedding. While they are similar in that they also provide explanations for each dimension, the explanations are in the form of an interpretability score that measures how well a dimension is associated with a subgroup of nodes. ~\cite{khoshraftar2021centrality} define the interpretability of an embedding algorithm as the extent to which its embedding dimensions can represent "important" nodes in a category with extreme values, where "importance" is based on centrality measures. Carrying over work from traditional explanation methods like SHAP and LIME, \cite{gui2022flowx} propose FlowX, a method of explaining GNNs by identifying important message flows using Shapley values. A flow here is defined as the set of $T$ messages that are passed in a message passing network with $T$ layers. Each flow for a given node is then treated as a player for the Shapley formulation. \cite{huang2022graphlime} propose GraphLIME that learns an interpretable model locally in the subgraph of the node being explained. As discussed earlier, \cite{yuan2021explainability} use Monte Carlo tree search to explore subgraphs, but use Shapley values to compute subgraph importance. \cite{duval2021graphsvx} propose GraphSVX that captures the contribution of each feature and node towards the prediction by constructing a surrogate model on a perturbed dataset. Here, the explanations come in the form of Shapley values. 

In terms of objective functions, ~\cite{duong2019interpretable} propose a loss function that uses a notion of minimizing the number of connections between communities, and in doing so, the nodes in a community are embedded closer to each other. Explainability comes in the form of each dimension defining the degree of membership to a community. ~\cite{rossi2018deep} use features akin to what we refer to as sense features as base features, to which relational feature operators are applied to generate embeddings. Explainability comes in the form of examining which relational operators were used to generate the embeddings.  

Compared with related work in the field, our proposed method \textsc{XM} is independent of any particular downstream task, can be used to extend any existing embedding algorithm, and works with any set of sense features. It also provides explanations on the node- and dimension-level.

\hide{
\section{Future Work }
\label{conclusion_future_section}
The low level granularity of our explanations could be used as building blocks to generate subgraph or graph level explanations by using a diffusion process to diffuse the explanation at each node as proposed by ~\cite{rossi2018deep}. In future work, we intend to compare the explanations that this generates to the current state-of-the-art explanation techniques for graph-level explanations discussed earlier. 

We also notice that the Explain matrices tend to have dimensions that do not contribute to explaining any sense features, i.e. the rows corresponding to these dimensions in the Explain matrix are near zero. If similar behavior is observed when looking at subgraph- or graph-level explanations, we could explore the merit of using this information to guide model selection by investigating whether we can reduce the number of embedding dimensions if a dimension does not contribute to the explanation of the graph.
}

\section{Conclusion and Future Work}
We present a method for generating explanations for each node in the form of an Explain matrix using a set of human-understandable ``sense'' features. These features can be based on the graph structure (such as degree, clustering coefficient, and PageRank) or other user-defined features. We use the nuclear norm of the generated Explain matrices to quantify their usefulness. The choice of nuclear norm is due to its relationship with the entropy of a matrix. In particular, reducing the nuclear norm leads to the reduction of the lower bound of the entropy of the Explain matrices. Moreover, we introduce the \textsc{XM} framework, which modifies existing embedding algorithms to produce embeddings with low nuclear norms on Explain matrices, while maintaining downstream task performance. We demonstrate the effectiveness of \textsc{XM} across 4 embedding algorithms and 6 real-world datasets and analyze the impact of \textsc{XM}'s constraints through an ablation study. \textsc{XM} is independent of any downstream task and can be used with any existing embedding algorithm and any set of sense features. 

\paragraph{Future work.} We  noticed that the Explain matrices tend to have dimensions that do not contribute to explaining any sense features -- i.e. the rows corresponding to these dimensions in the Explain matrix are near zero. Such information can be used for model selection -- i.e. to set the hyperparameter associated with the number of embedding dimensions.


\bibliography{main}
\bibliographystyle{tmlr}

\appendix
\section{Appendix}

\subsection{Variations to the Objective Function}

Recall that we define the Explain matrix for a node $k$:
\begin{align}
    E_k = \frac{\vec{y_k} \otimes\vec{f_k}^T}{\|\vec{y_k}\|\|\vec{f_k}\|}
\end{align}

Thus, the sparsity and orthogonality constraints (as defined in Equations (\ref{sparse_eq}) and (\ref{ortho_eq}), respectively) are on nodes. To add pairwise constraints on nodes, we investigate variations to Equation (\ref{ortho_eq}). 

Specifically, we implement the following variations for any node pair $i$ and $j$ : 

\begin{enumerate}
    \item Minimize $a_{ij} \|E_i - E_j\|_F$. \\
    This constrains the $E$ matrices of each pair of nodes $i, j$ to be similar if they are connected by an edge (i.e. if $a_{ij} = 1$), where the similarity is captured in terms of the Frobenius norm of the difference between their two Explain matrices.

    \item Minimize $\|E_iE_j^T - f_if_j^T\|_F$.\\ 
    Here, we move away from asking whether or not two nodes are connected by an edge. We compute the dimension-wise similarities between the Explain matrices and the similarities between sense feature vectors for all possible node pairs and obtain two distance matrices, each of dimension $\mathbb{R}^{n \times n}$. We then minimize the Frobenius norm of the difference between these two matrices. 
    
\end{enumerate}

We find that adding pairwise constraints to the objective function is not a good idea. The outcome is very sensitive to the choice of hyperparameters and the reduction in nuclear norm is not as significant. We posit that this may be due to opposing components of the objective function. For example, consider a high-degree node (a.k.a.~a hub) that is connected to a low-degree node. In this case, most  algorithms place the embeddings of the two nodes close to each other (since the nodes are connected) even though their sense features are different. To summarize, we find that using the formulation defined in Equations (\ref{sparse_eq}) and (\ref{ortho_eq}) leads to the most stable setup with the largest reduction in nuclear norms among the ones described above.

\subsection{Nuclear Norm Distributions for Explain Matrices}
\label{nuclear_norm_distribution_section}
\begin{figure}[h]
    \begin{center}
        \includegraphics[width=1.0\linewidth]{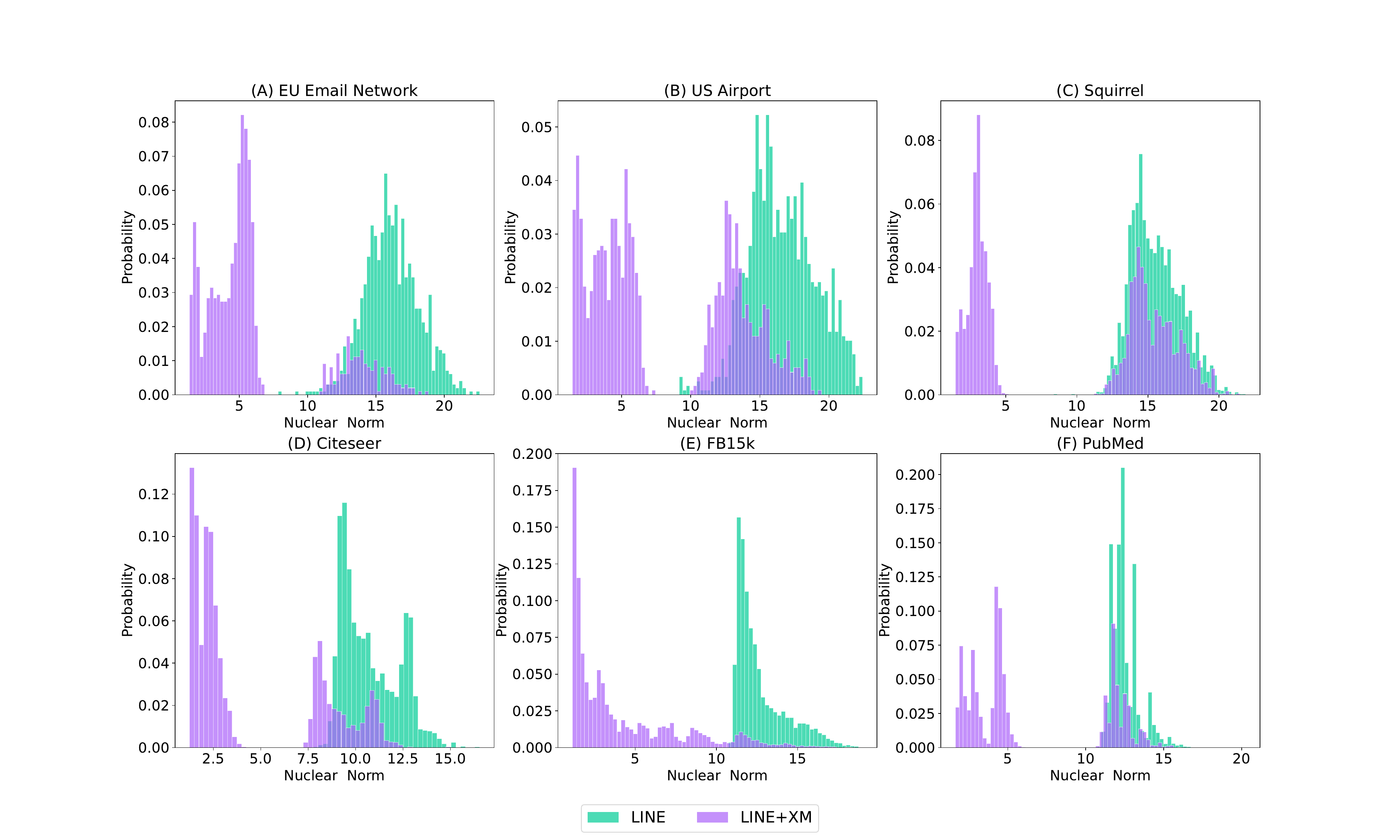}
    \end{center}
    
    \caption{{Nuclear norm distribution across the Explain matrices for LINE and LINE+XM. Observe how the distribution of LINE+XM is shifted to the left with a lower mean across each of the 6 datasets.  
    }}
    \label{fig:figure_8}
\end{figure}

\begin{figure}[h]
    \begin{center}
        \includegraphics[width=1.0\linewidth]{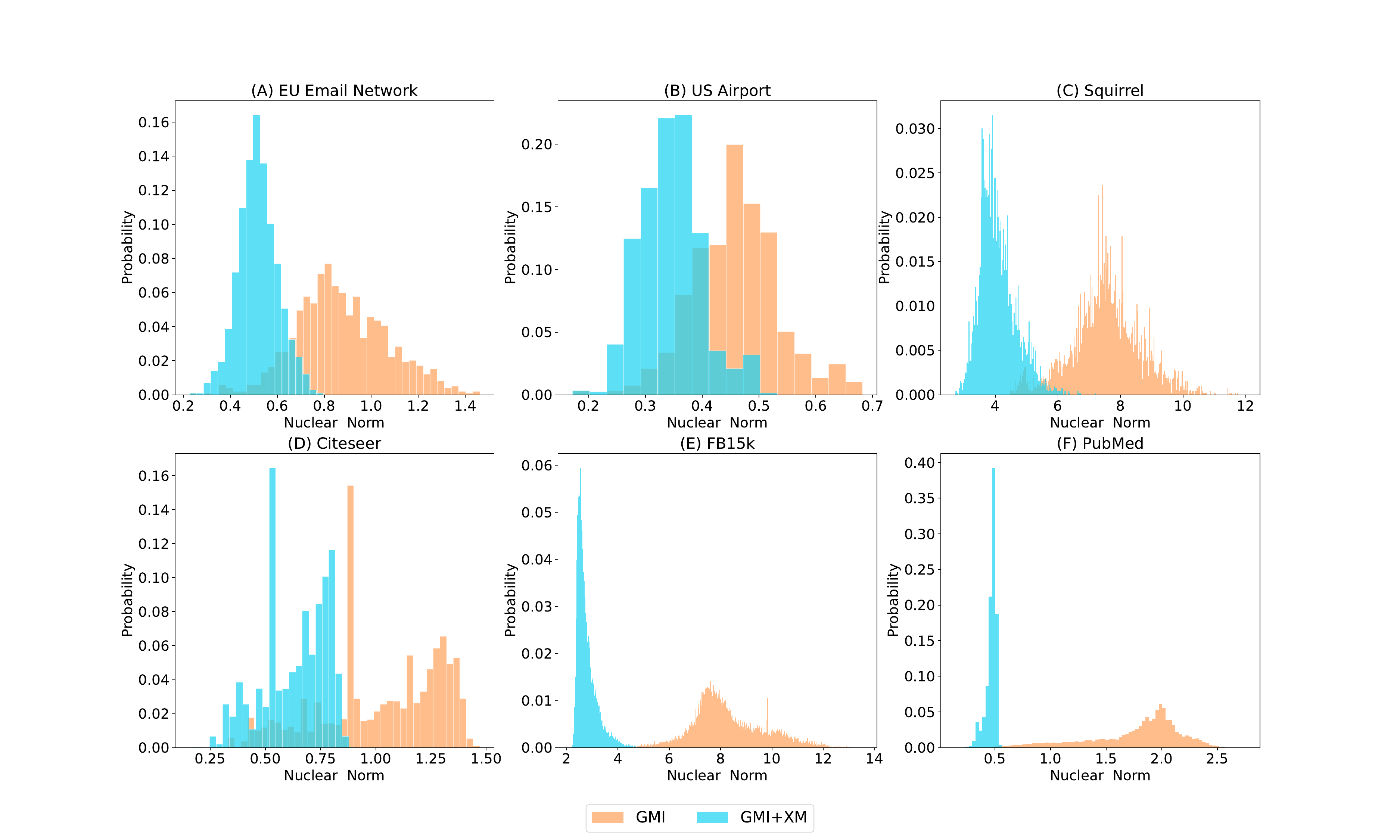}
    \end{center}
    
    \caption{{Nuclear norm distribution across the Explain matrices for DGI and DGI+XM. Observe how the distribution of DGI+XM is shifted to the left with a lower mean across each of the 6 datasets.  
    }}
    \label{fig:figure_9}
\end{figure}

\begin{figure}[h]
    \begin{center}
        \includegraphics[width=1.0\linewidth]{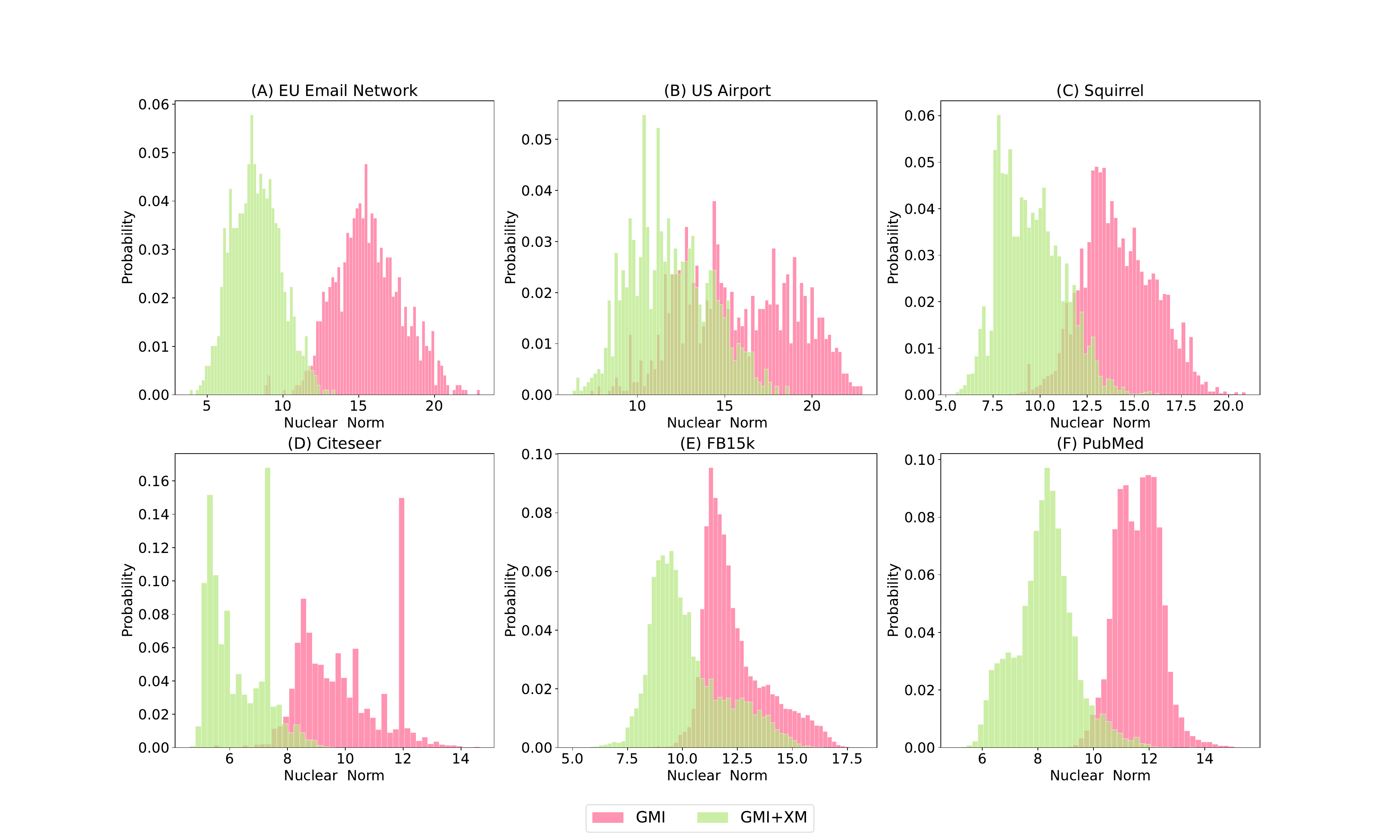}
    \end{center}
    
    \caption{{Nuclear norm distribution across the Explain matrices for GMI and GMI+XM. Observe how the distribution of GMI+XM is shifted to the left with a lower mean across each of the 6 datasets.  
    }}
    \label{fig:figure_10}
\end{figure}

Recall that each node has a corresponding Explain matrix. Thus, for each algorithm and dataset, we calculate the nuclear norm distribution across the Explain matrices of nodes.  Figures~\ref{fig:figure_8},~\ref{fig:figure_9}, and~\ref{fig:figure_10}, respectively, show the nuclear norm distributions between LINE and LINE+XM, DGI and DGI+XM, and GMI and GMI+XM for all 6 datasets. We observe that the \textsc{XM} variants produce Explain matrices whose nuclear norms are lower than the original versions.

\subsection{Runtime Analysis}
\label{runtime_analysis_section}
\begin{figure}[h]
    \begin{center}
        \includegraphics[width=0.6\linewidth]{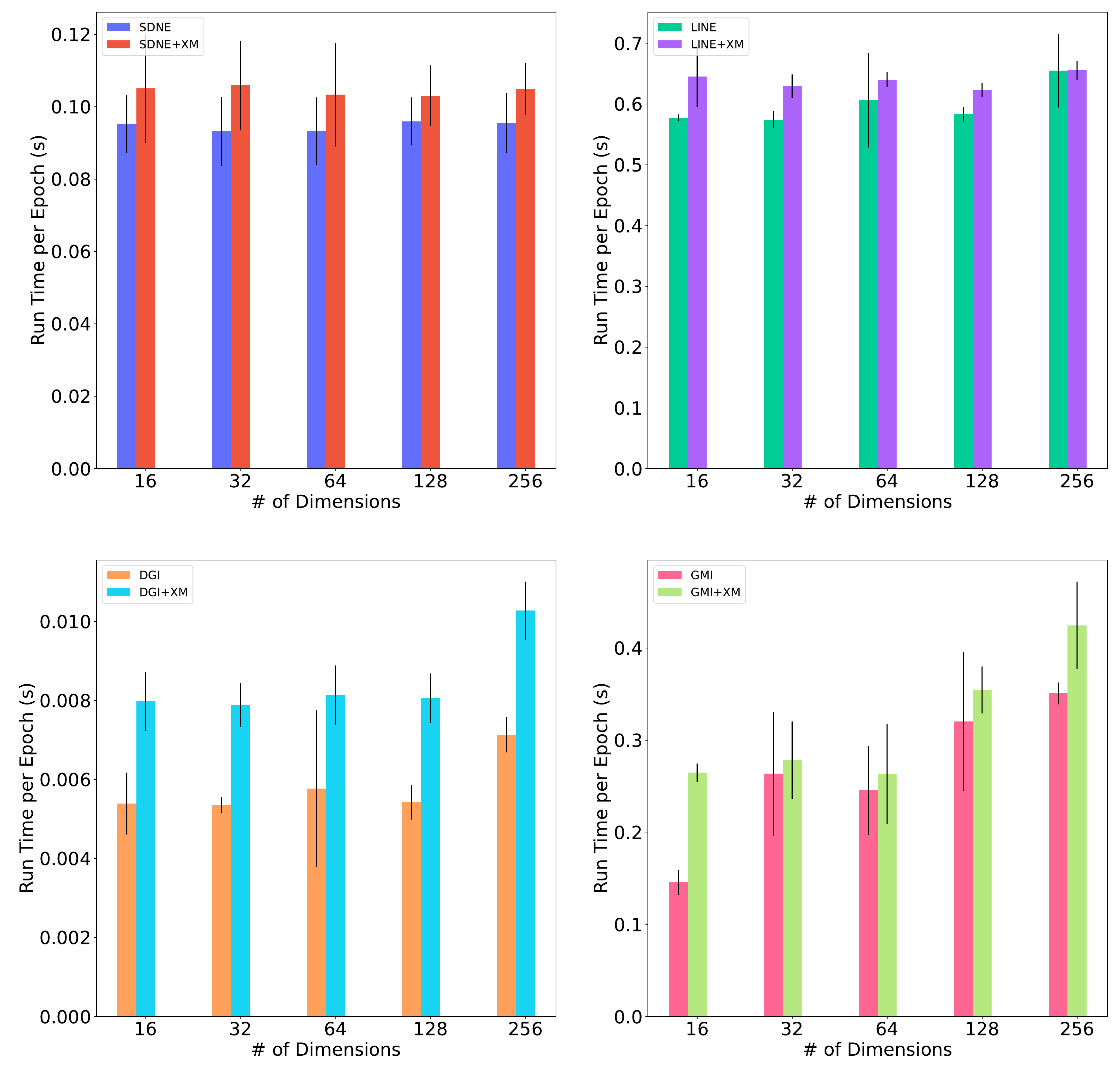}
    \end{center}
    
    \caption{{US Airport network: runtimes per epoch (in seconds) for SDNE, LINE, DGI, GMI, and their corresponding \textsc{XM} variants. Results are averaged across 5 runs. Observe that the runtimes per epoch are comparable to the original versions. Error bars show standard error. }}
    \label{fig:figure_11}
\end{figure}

\begin{figure}[h]
    \begin{center}
        \includegraphics[width=0.5\linewidth]{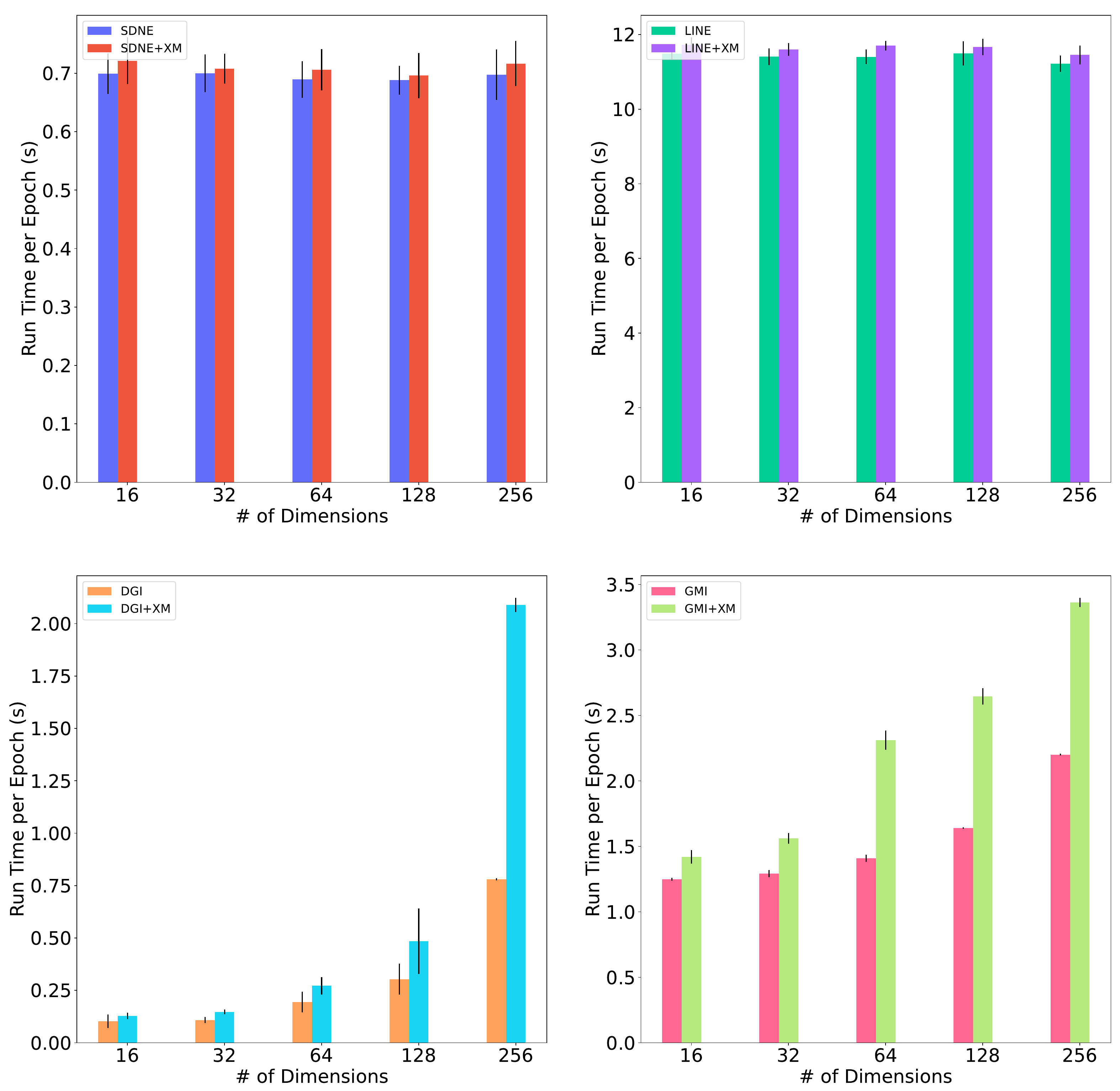}
    \end{center}
    
    \caption{{Squirrel network: runtimes per epoch (in seconds) for SDNE, LINE, DGI, GMI, and their corresponding \textsc{XM} variants. Results are averaged across 5 runs. Observe that the runtimes per epoch are comparable to the original versions. Error bars show standard error. }}
    \label{fig:figure_12}
\end{figure}

\begin{figure}[h]
    \begin{center}
        \includegraphics[width=0.5\linewidth]{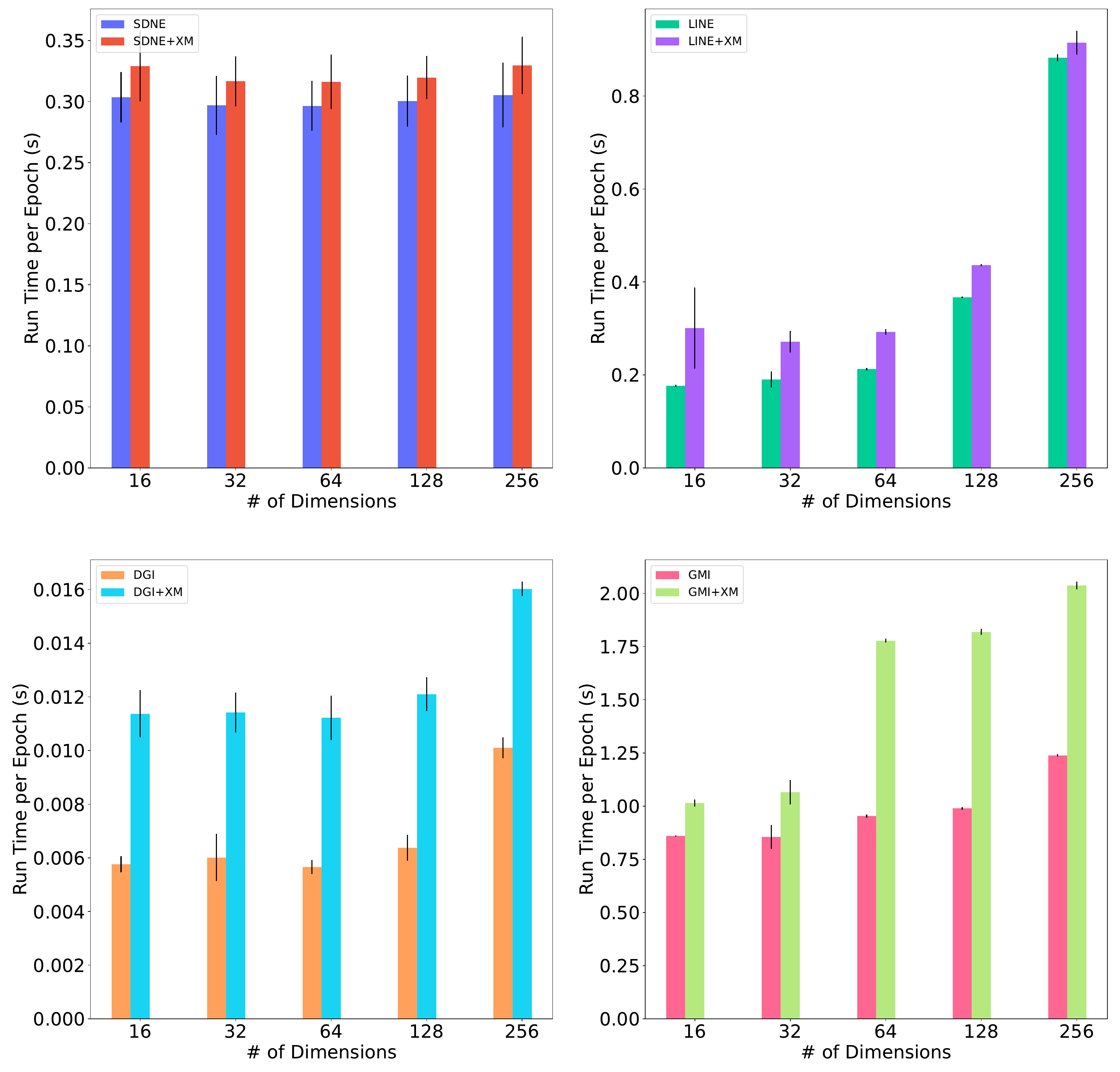}
    \end{center}
    
    \caption{{Citeseer network: runtimes per epoch (in seconds) for SDNE, LINE, DGI, GMI, and their corresponding \textsc{XM} variants. Results are averaged across 5 runs. Observe that the runtimes per epoch are comparable to the original versions. Error bars show standard error. }}
    \label{fig:figure_13}
\end{figure}

\begin{figure}[h]
    \begin{center}
        \includegraphics[width=0.5\linewidth]{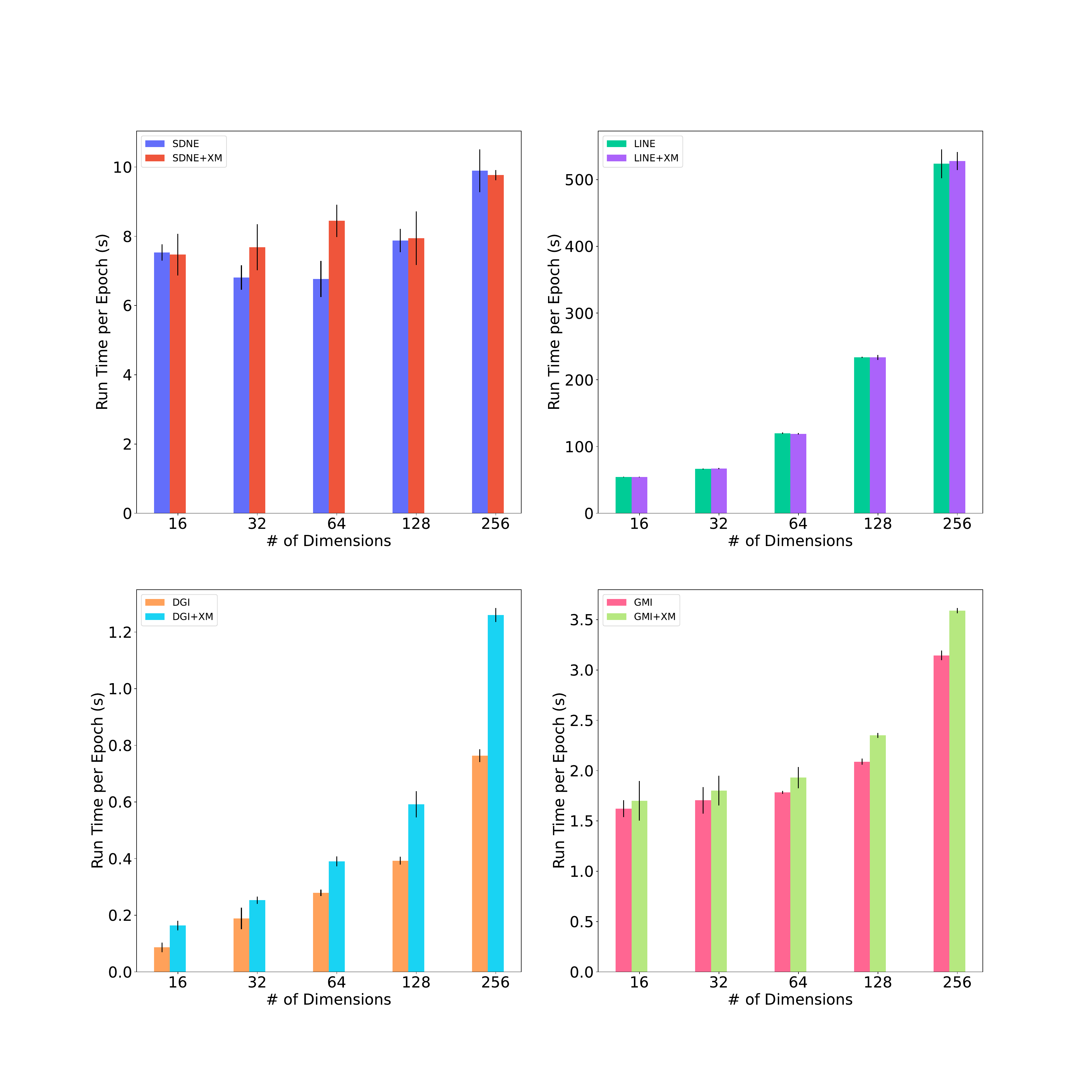}
    \end{center}
    
    \caption{{FB15k network: runtimes per epoch (in seconds) for SDNE, LINE, DGI, GMI, and their corresponding \textsc{XM} variants. Results are averaged across 5 runs. Observe that the runtimes per epoch are comparable to the original versions. Error bars show standard error. }}
    \label{fig:figure_14}
\end{figure}

\begin{figure}[h]
    \begin{center}
        \includegraphics[width=0.5\linewidth]{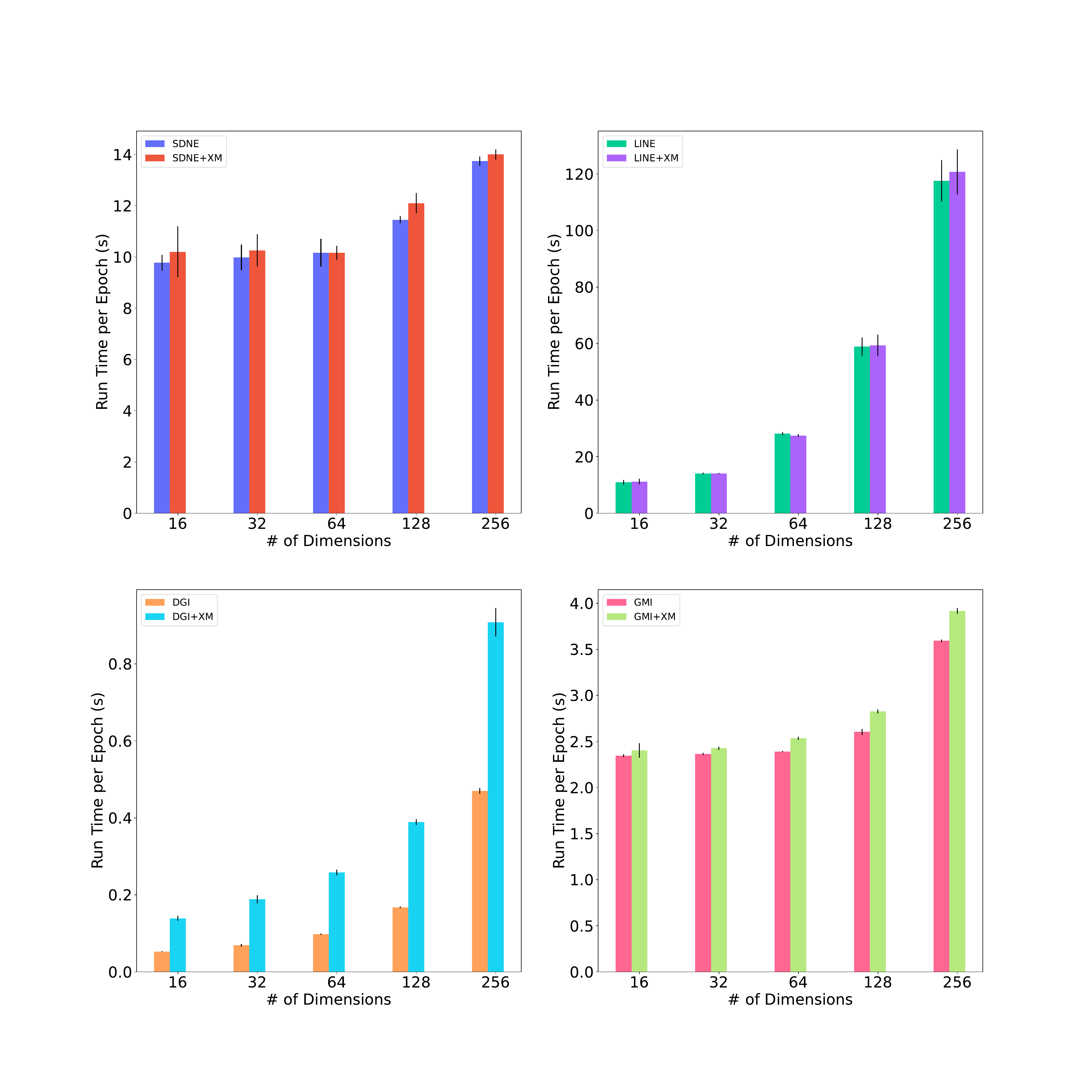}
    \end{center}
    
    \caption{{PubMed network: runtimes per epoch (in seconds) for SDNE, LINE, DGI, GMI, and their corresponding \textsc{XM} variants. Results are averaged across 5 runs. Observe that the runtimes per epoch are comparable to the original versions. Error bars show standard error. }}
    \label{fig:figure_15}
\end{figure}

We examine differences in run time per epoch (in seconds) between the original algorithms and the \textsc{XM} variants across different embedding dimensions. 
Figure \ref{fig:figure_11} shows the run time comparison for the US Airport network, Figure \ref{fig:figure_12} for the Squirrel network, Figure \ref{fig:figure_13} for the Citeseer network, Figure \ref{fig:figure_14} for the FB15k network, and Figure \ref{fig:figure_15} for the PubMed network.
We observe that the \textsc{XM} variants are comparable in runtime to their original versions.

\subsection{Nuclear Norm Minimization}
\begin{figure}[!htb]
    \centering
    \subfloat[{We average the nuclear norms of the Explain matrices of each node in the network. The experiment is repeated 5 times and average results are shown. Error bars depict standard error. Observe that the mean nuclear norms for the \textsc{XM} variants are lower than the original versions. The differences in means are statistically significant with $p$-values less than 0.05.}]{
        \includegraphics[width=0.8\textwidth]{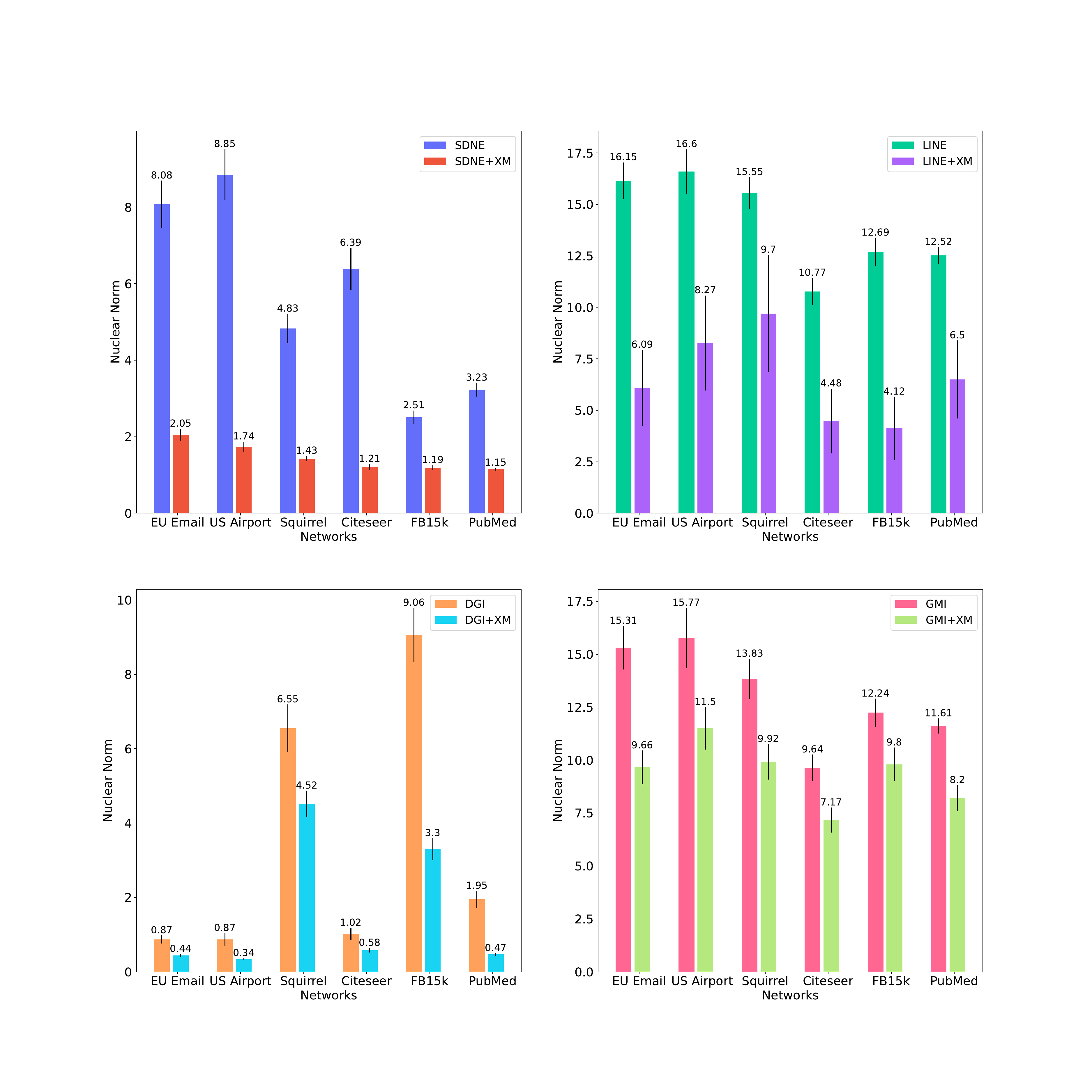}
        \label{fig:figure_16}
    }
    
    \subfloat[$p$-values associated with Figure~\ref{fig:figure_16}. The nuclear norms between the original versions and the \textsc{XM} variants are statistically significant.]{
        \begin{tabular}{|c|c|c|c|c|c|c|}
        \hline
                & EU Email & US Airport & Squirrel & Citeseer & FB15k & PubMed \\ \hline
            SDNE vs.~SDNE + XM & 2.13e-10 & 9.78e-11 & 1.65e-10 & 7.76e-10 & 4.44e-11 & 3.96e-11 \\ \hline
            LINE vs.~LINE + XM & 1.96e-07 & 8.86e-06 & 6.72e-05 & 8.30e-06 & 1.38e-07 & 3.92e-06 \\ \hline
            DGI vs.~DGI + XM   & 1.62e-03 & 1.47e-04 & 2.62e-04 & 4.88e-04 & 1.93e-07 & 5.03e-07 \\ \hline
            GMI vs.~GMI + XM   & 1.80e-06 & 6.59e-06 & 3.81e-06 & 1.02e-07 & 2.88e-05 & 1.60e-07 \\ \hline
        \end{tabular}
        \label{table:p_values}
    }
    
    \caption{Quantitative evaluation of the Explain matrices based on their nuclear norms across networks, original methods and their \textsc{XM} variants. The lower the norm, the better the Explain matrix.} 
    \label{fig:combined}
\end{figure}

Figure \ref{fig:figure_16} looks at all networks in Table \ref{table:real_nets} embedded using each of the 4 algorithms from Table \ref{table:algo_table}. Results are averaged across 5 runs. We see that the nuclear norms of the Explain matrices are lower for the \textsc{XM} variants across each network. These differences are statistically significant with $p$-values less than 0.05. Table~\ref{table:p_values} lists the $p$-values.

\end{document}